\documentclass[final,5p,times,twocolumn, sort&compress,authoryear]{elsarticle}

\usepackage{amssymb}
\usepackage{amsmath}

\usepackage{hyperref}
\usepackage{hyperref}
\usepackage{graphicx,verbatim}
\usepackage{amsmath}
\usepackage{amssymb}
\usepackage{mathtools}
\usepackage{amsthm}
\usepackage[capitalize,noabbrev]{cleveref}
\usepackage{amsmath, amssymb}

\usepackage{multirow}
\usepackage{booktabs}
\usepackage[table,xcdraw]{xcolor} 
\usepackage{verbatim}
\usepackage{xcolor}
\definecolor{lightgreen}{RGB}{204, 255, 204}
\usepackage{amsmath}
\usepackage{subcaption}
\usepackage{lscape}    
\usepackage{amsthm}
\usepackage{booktabs}
\usepackage{enumitem}
\usepackage{algorithm}
\usepackage{algpseudocode}

\algnewcommand\Input{\item[\textbf{Input:}]}%
\algnewcommand\Output{\item[\textbf{Output:}]}%

\algblockdefx{Indent}{EndIndent}{}{}          
\algrenewcommand\algorithmicindent{1.5em}     
\algnotext{Indent}\algnotext{EndIndent} 

\usepackage{natbib}

\newcommand{\bd}{\boldsymbol} 
\newcommand{\mb}{\mathbf} 
\newcommand{\be}{\begin{equation}}
\newcommand{\ee}{\end{equation}}
\usepackage{graphicx}  
\usepackage{subcaption}

\definecolor{splatcolor}{RGB}{70,130,180}   
\definecolor{blurcolor}{RGB}{60,179,113}    
\definecolor{slicecolor}{RGB}{205,92,92}    

\makeatletter
\renewcommand{\fnum@algorithm}{\fname@algorithm~\thealgorithm.}
\makeatother

\DeclareMathOperator{\diag}{diag}

\journal{arXiv}

\begin{document}

\begin{frontmatter}





\title{Unleashing Diffusion and State Space Models for Medical Image Segmentation}

\author[label1]{Rong Wu} 
\author[label2]{Ziqi Chen} 
\author[label3]{Liming Zhong} 
\author[label4]{Heng Li}

\author[label1]{Hai Shu\corref{cor1}}

\affiliation[label1]{organization={Department of Biostatistics, School of Global Public Health, New York University},city={New York},state={NY},country={USA}}

\affiliation[label2]{organization={School of Statistics, KLATASDS-MOE, East China Normal University, Shanghai, China}}

\affiliation[label3]{organization={School of Biomedical Engineering, Southern Medical University, Guangzhou, China}}

\affiliation[label4]{organization={Faculty of Biomedical Engineering, Shenzhen University of Advanced Technology, Guangzhou, China}}

\cortext[cor1]{Corresponding author: hs120@nyu.edu}

\begin{abstract}
Existing segmentation models trained on a single medical imaging dataset often lack robustness when encountering unseen organs or tumors. Developing a robust model capable of identifying rare or novel tumor categories not present during training is crucial for advancing medical imaging applications. 
We propose DSM, a novel framework that leverages diffusion and state space models to segment unseen tumor categories beyond the training data. DSM utilizes two sets of object queries trained within modified attention decoders to enhance classification accuracy. Initially, the model learns organ queries using an object-aware feature grouping strategy to capture organ-level visual features. 
It then refines tumor queries by focusing on diffusion-based visual prompts, enabling precise segmentation of previously unseen tumors. 
Furthermore, we incorporate diffusion-guided feature fusion to improve semantic segmentation performance. By integrating CLIP text embeddings, DSM captures category-sensitive classes to improve linguistic transfer knowledge, thereby enhancing the model's robustness across diverse scenarios and multi-label tasks. 
Extensive experiments demonstrate the superior performance of DSM in various tumor segmentation tasks. Code is available at  \url{https://github.com/Rows21/k-Means_Mask_Mamba}.
\end{abstract}


\end{frontmatter}



\section{Introduction}
A significant challenge in medical image segmentation lies in adapting a model to diverse segmentation tasks, particularly those involving rare diseases \citep{challenges}.
Most existing models, trained on a single medical imaging dataset, lack reliability and robustness when applied to different organs or tumors.
Recent studies \citep{clip-driven, jiang2024zept} have aimed to develop a unified model capable of addressing a broad spectrum of organs and tumors.
However, these models often struggle with blurred edge detection in multi-class object detection and segmentation tasks, and continue to face difficulties in identifying clinically relevant rare or novel lesion classes.
For many types of cancer, the ability to accurately detect  and  segment tumors is critical, directly impacting treatment decisions and cancer prognosis.
In such scenarios, there is a strong demand for a medical image segmentation model that can effectively identify rare diseases with limited instances and  precisely segment tumor lesions.
In this study, we integrate Diffusion Models, State Space Models (SSMs), and Open-Vocabulary Semantic Segmentation (OVSS) to further enhance performance across various tumor segmentation tasks.

SSMs \citep{gu2021efficiently, gu2023mamba} have emerged as a prominent topic in the fields of computer vision and language modeling. To address the challenges associated with long sequence modeling, Mamba~\citep{gu2023mamba}, derived from the foundational SSM framework  \citep{hamilton1994state}, offers a solution designed to capture extensive dependencies and enhance training efficiency via a selection mechanism.
Numerous studies have explored the application of Mamba across various imaging tasks.
U-Mamba \citep{ma2024u} integrates the Mamba layer into the encoder of nnUNet \citep{isensee2021nnu} to boost general medical image segmentation capabilities.
 Vision Mamba \citep{zhu2024vision} introduces a bidirectional SSM for data-driven global visual context modeling and incorporates position embeddings for spatially informed visual comprehension.
Mamba-2 \citep{dao2024transformers} highlights the significant performance gains achieved by effectively integrating the Transformer module 
\citep{vaswani2017attention} with SSM within the network.
Nonetheless, the fusion of Transformer and SSM remains underexplored, particularly in the domain of medical image segmentation tasks.

Diffusion models \citep{peebles2023scalable,wu2024medsegdiff} have recently garnered significant attention for their utility in addressing medical imaging challenges. 
A key advantage of diffusion models is their ability  to generate high-quality samples comparable to those produced by Generative Adversarial Networks (GANs), while demonstrating superior robustness against mode collapse \citep{wu2024medsegdiff, wang2023dformer}.
Initially, diffusion models were developed to  augment datasets and harmonize sample distributions. 
However, recent studies have highlighted their effectiveness in addressing issues related to visual boundary enhancement in imaging tasks \citep{tan2022semantic, zhang2024tale}.

OVSS methods \citep{OVSeg, freeseg}, incorporating vision-language models (VLMs) like CLIP~\citep{clip}, have found diverse applications within the broader image processing domain. 
This methodology facilitates the transfer of knowledge from fixed categories to unknown (novel) ones by learning transferable representations from the annotated data of known classes to represent unknown ones \citep{bert, joulin2016bag}. 
Leveraging the robust zero-shot transfer capabilities demonstrated by CLIP, researchers have been actively developing advanced open-vocabulary models by harnessing extensive image-text pairs for detailed vision tasks in detection \citep{zareian2021open, clip-driven} and segmentation \citep{huynh2022open, rao2022denseclip}. 
Some strategies integrate object queries from MaskFormer  models \citep{maskformer, mask2former}, trained on base categories, to produce class-agnostic mask proposals. 
These proposals are subsequently aligned with tokens created by VLMs, showcasing remarkable zero-shot segmentation prowess.

In  medical imaging, developing  a unified model  has the potential to streamline data integration by amalgamating information from various medical institutions. This amalgamation allows the model to acquire knowledge from a broader spectrum of data, thereby enhancing its adaptability and learning capacity.
In clinical settings, the use of unified models eliminates the need for technicians to switch between multiple models while handling various tasks.
Nevertheless, these models require substantial amounts of labeled data and face challenges in identifying rare or novel lesion categories that are clinically important.

To tackle these challenges, we introduce a novel framework,  $\mb{D}$iffusion and $\mb{S}$tate Spaces $\mb{M}$odel (DSM), designed to: 
(1) enhance semantic segmentation between organs and tumors, and (2) boost the robustness of multi-organ recognition capabilities.
Our DSM involves setting up two types of queries—organ queries and tumor queries—to represent distinct embeddings. The organ queries are trained during a preliminary stage.
Subsequently, DSM integrates multi-scale feature maps 
with diffusion-guided refinement maps
to improve tumor boundary detection.
Following this, DSM creates mask prompts for out-of-distribution  regions within organs to help accurately identify tumor features.
The refined mask features and mask prompts are then fed into Transformer decoders to detect tumor region, facilitating accurate segmentation and recognition tasks.

Our main contributions are summarized as follows:
\begin{itemize}
\item DSM incorporates a novel k-Means Mask Mamba (kMMM) layer, a key innovation designed to enhance the model’s ability to retain and utilize long-term memory for query embeddings. By effectively modeling long-range dependencies, the kMMM layer ensures that crucial information is preserved during the processing of query embeddings, leading to more accurate and robust predictions.

\item DSM addresses the challenge of distinguishing between tumor and organ regions by introducing a mask prompt after the vision decoder. This prompt prioritizes the detection of tumor features within organ regions by focusing on the model’s attention on out-of-distribution  regions. This approach not only improves the model’s ability to detect and segment tumors but also ensures robustness across diverse datasets.

\item DSM leverages a diffusion-guided query decoder to refine the segmentation process by focusing on the precise delineation of tumor boundaries. This decoder integrates diffusion processes to smooth and enhance feature maps, enabling the capture of subtle variations in tumor boundaries
and improving overall segmentation accuracy.
\end{itemize}

\section{Related Work}
\subsection{Multi-Organ and Tumor Segmentation}
Advancements in U-Net-based architectures \citep{fang2020multi, isensee2021nnu} and deep learning methodologies \citep{tang2022self} have significantly propelled the field of multi-organ and tumor segmentation.
Since the introduction of the Segment Anything Model (SAM) \citep{sam}, the integration of subsequent transformer blocks into the network architecture backbone \citep{ma-sam, gu2020context} has been under investigation.
Our study primarily delves into addressing early-stage tumor detection issues in medical image segmentation, where current solutions exhibit limited performance, particularly in multi-organ segmentation \citep{clip-driven,wu2024medsegdiff}. Hence, we explore a novel architecture that combines Transformer \citep{vaswani2017attention} and Mamba \citep{gu2023mamba} models to enhance segmentation performance in clinical scenarios, with  emphasis on detecting and diagnosing minority tumors \citep{moor2023foundation}.

\subsection{Open-Vocabulary Semantic Segmentation}
The rapid progress in large-scale language models, such as GPT \citep{brown2020language}, BERT \citep{bert}, and Med-PaLM \citep{singhal2023large}, has significantly advanced natural language understanding and generation. These developments have inspired the emergence of large-scale pre-trained vision-language models (VLMs) that integrate visual and textual information for more generalizable and context-aware perception \citep{li2022language, cris, per-clip}. In particular, open-vocabulary semantic segmentation (OVSS),  where models are tasked with segmenting arbitrary objects based on natural language descriptions rather than fixed class labels, has become a promising paradigm for addressing the limitations of closed-set segmentation systems.

\subsection{Prompt-based Zero-shot Models}
Recent work on large language models has explored  in-context learning for medical image segmentation and can segment any class guided by example images or text \citep{wang2025sam,butoi2023universeg}. SAMMed \citep{wang2025sam} fine-tunes SAM on large 3D medical datasets using the bounding box but lacks the ability for detailed editing. UniverSeg~\citep{butoi2023universeg} generalizes well across diverse medical imaging tasks and modalities by leveraging a unified architecture trained on multiple datasets with task-specific prompts. However, it struggles with unseen tumors whose associated organs or rare anatomical structures are not well represented in  training data, and its performance depends heavily on the quality of prompt images.

\subsection{Unseen Tumor Detection and Localization}
Recent studies \citep{zegformer, yuan2023devil} have proposed novel approaches for detecting or segmenting previously unseen tumors or lesions, often framing the task as a form of out-of-distribution (OOD) detection. These methods aim to identify and localize anomalies that were not explicitly present in the training data, leveraging techniques such as distribution modeling or self-supervised representation learning. In the context of medical imaging, such unseen tumors often represent rare pathologies, subtle lesions, or atypical anatomical variations that deviate from the model's learned distribution.
It is essential to distinguish between traditional OOD detection and the zero-shot segmentation task explored in this paper. This distinction underlines the importance of open-vocabulary capabilities and cross-modal understanding in advancing the generalizability of segmentation models in medical imaging.

\section{Preliminaries}
\subsection{State Space Models}
SSMs are inspired by the continuous system that maps a 1-D function or sequence $x(t) \in \mathbb{R}$ to an output $y(t) \in \mathbb{R}$ through a hidden state $h(t) \in \mathbb{R}^N$.
This system uses evolution parameter $\mb{A} \in \mathbb{R}^{N \times N}$, projection parameters $\mb{B} \in \mathbb{R}^{N \times 1}$ and $\mb{C} \in \mathbb{R}^{1 \times N}$, and can be represented as the following linear ordinary differential equation:
\begin{align}
    &h'(t)=\mb{A}h(t) + \mb{B}x(t), \\
    &y(t) = \mb{C}h(t).
\end{align}
The Structured State Space sequence model (S4) \citep{gu2021efficiently}  and Mamba \citep{gu2023mamba} are discrete versions of this continuous system, which incorporate a step size $\Delta$ to approximate the continuous parameters $\{\mb{A}, \mb{B}\}$ into discrete approximations $\{\mb{\Bar{A}}, \mb{\Bar{B}}\}$. 
For example, Mamba uses the zero-order hold (ZOH)
discretization rule:
\begin{align}
    &\mb{\Bar{A}} = \exp{(\Delta\mb{A})},\\
    &\mb{\Bar{B}} = (\Delta\mb{A})^{-1}(\exp{(\Delta\mb{A})}-\mb{I})\Delta\mb{B}.
\end{align}
The discretized version of the system is then written as:
\begin{align}
    &h_t = \mb{\Bar{A}}h_{t-1} + \mb{\Bar{B}}x_t,\\
    &y_t = \mb{C}h_t,
\end{align}
where $x_t=x(t\Delta)$ and $h_{-1}=0$.
Thus, the models can be computed via a global convolution.

\subsection{Diffusion Process}\label{diff}
Diffusion process \citep{perona1990scale, tan2022semantic} is capable of learning a sequence of state transitions to generate high-quality samples from noise. Other than the generation capabilities derived from noise in diffusion models, we leverage the diffusion process in this work to improve visual boundary clarity, thereby enhancing segmentation performance.
This process involves designing a diffusivity function $D$ to solve a second-order partial differential equation.
Given a feature map $\mb{F}$ to be smoothed, the diffusion process is described~as:
\begin{align}
    \frac{\partial \mb{F}}{\partial t} = Div(D(\mathbf{F})\cdot \nabla \mb{F}), \label{eqdiff}
\end{align}
where $\frac{\partial \mb{F}}{\partial t}$ is the time derivative of the solution matrix, $\nabla $ is the gradient operator, $Div$ is the divergence operator, and $D$ is a diffusivity function depending on the input.  
As time $t$ progresses, the solution $\mathbf{F}$ gradually becomes  smoother compared to the original matrix, as discussed in  \cite{tan2022semantic}.
In numerous medical image segmentation scenarios, the blurring of class boundaries is a common concern.
Our study aims to extract refined boundary features by employing linear diffusion processes that smooth both backgrounds and edges. 
This diffusion process applies more smoothing to regions parallel to boundaries than to regions perpendicular to these edges. Specifically, given a feature $\mb{F}$ in areas with ambiguous boundaries, it undergoes a time-stepwise diffusion process. The intention is for the final state of the diffused feature to precisely delineate the boundaries between different anatomical structures.

\section{Method}

Our DSM framework operates in two stages: Stage~1 focuses on organ queries generation, and Stage~2 on tumor queries refinement. This section elaborates on both stages.

\subsection{Stage 1: Organ Queries Generation}
Our Stage 1 network structure is constructed based on  MaskFormer \citep{maskformer, mask2former}. A vision encoder extracts pixel features using either a CNN \citep{he2016deep} or a Transformer \citep{liu2021swin} backbone, while the vision decoder is  responsible for reconstructing multi-scale feature maps $\{\mathbf{F}_i\}_{i=1}^4$, where $\mathbf{F}_i \in \mathbb{R}^{H_iW_iD_i \times C_i}$, with $H_i$, $W_i$, $D_i$, and $C_i$ denoting the height, width, depth, and channel dimensions of feature maps for each decoder layer, respectively. Finally, the vision decoder within the backbone network progressively up-samples visual features into high-resolution image embedding $\hat{\mathbf{F}}_4 \in \mathbb{R}^{HWD \times C}$.

In Stage 1, a set of organ queries $\mathbf{O} \in \mathbb{R}^{N_o \times C}$ is randomly generated in the model, where $N_o$ is the number of  organ types. A k-Means Mask Mamba (kMMM) decoder $K_d$ updates the queries by interacting with multi-scale visual feature maps $\{\mb{F}_i\}_{i=1}^4$. To capture organ-level information and tackle challenges related to long-tailed distributions, a proposed object-aware feature grouping strategy is implemented in $K_d$. Additionally, to enhance global context understanding in query features, a SSM layer is employed to maintain long-term knowledge in the kMMM decoder. This approach guides each organ query to accurately represent and specify its corresponding organ category.

\subsubsection{k-Means Mask Mamba}
\begin{figure}
	\centering
        \includegraphics[width=0.4\textwidth]{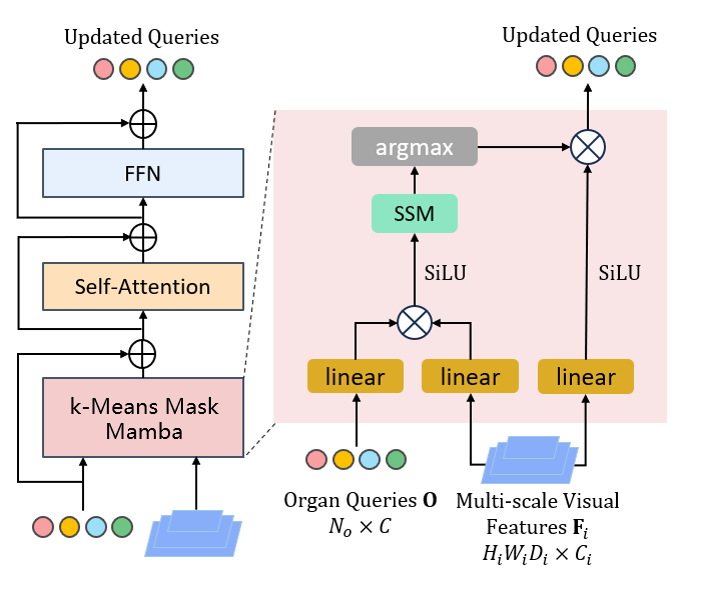}
	\caption{The k-Means Mask Mamba (kMMM) structure: To convert a Transformer decoder into a kMMM decoder, we add an SSM block after the cross-attention between queries and features.}
	\label{ssm}
\end{figure}

As shown in Figure \ref{ssm}, $K_d$ consists of a series of SSM and attention blocks that enable the queries to interact with multi-scale features. 
The traditional cross-attention layer updates the organ queries $\mb{O} \in \mathbb{R}^{N_o \times C}$ by
\begin{align}
    \hat{\mb{O}} = \mb{O} + \mathop{\arg\max}_{N_o} (\mb{Q}_O\mb{K}_F^T)\mb{V}_F,
\end{align}
where the superscripts $O$ and $F$ represent query and visual features, respectively.
As shown in Figure \ref{ssm}, the $\mathop{\arg\max}$ on $N_o$ acts as a query-wise grouping strategy as in k-means, attempting to group feature maps into query embedding.
We leverage this query-wise argmax from kMaX-DeepLab  \citep{yu2022k} to replace spatial-wise softmax in the initial cross-attention settings, aligning visual features with queries and mitigating the issue of long-tail problems.
Inspired by Mamba \citep{gu2021efficiently}, we employ a SSM layer before $\mathop{\arg\max}$ to maintain stable query embedding while training.
In the $i$-th kMMM block, the model leverages global information from multi-scale feature maps $\mathbf{F}_i$ through a k-means SSM layer, implemented as below:
\begin{align}
        &\mb{R}_i = \mathop{\arg\max}_{N_o} ({\rm{SSM}}(\mb{O}_i\mb{F}^T_{i})), \label{kmax1}
        \\ &\hat{\mb{O}}_i = \mb{R}_i  \mb{F}_i, \label{kmax2}
\end{align}
where $\mb{O}_i \in \mathbb{R}^{N_o \times C_i}$, $\mb{F}_i \in \mathbb{R}^{H_iW_iD_i \times C_i}$, and $\mb{R}_i \in \mathbb{R}^{N_o \times H_iW_iD_i}$ represent queries, feature maps, and query response, respectively. Query embedding $\hat{\mb{O}}_i$ is the linear projection from $\mb{O}$ for each decoder layer.
The features then go through SiLU activation function \citep{elfwing2018sigmoid} and the SSM layer to maintain long-term knowledge memory. Finally, we adopt a cluster-wise argmax to update the organ queries as $\hat{\mb{O}}_i$ and go through a projection layer to obtain $\hat{\mb{O}}\in \mathbb{R}^{N_o\times C}$.

As depicted in Figure \ref{overview}(a), after  processing through the vision decoder and kMMM decoder, we obtain updated organ queries $\hat{\mathbf{O}}$ and high-resolution image embeddings $\hat{\mathbf{F}}_4\in \mathbb{R}^{HWD\times C}$. The semantic  soft masks $\hat{\mathbf{Y}} \in [0, 1]^{N_o \times HWD}$ for organ queries are derived through a fixed match between the query embedding $\hat{\mathbf{O}}$ and high-resolution image features $\hat{\mathbf{F}}_4$, followed  by a sigmoid function. 

\subsubsection{Loss Functions}\label{loss}
We utilize the Dice loss to supervise the  predicted soft masks overall and a Binary Cross-Entropy (BCE) loss to supervise the category information of organs after processing query embeddings through an MLP layer.
Denote ground-truth one-hot labels $\mb{Y}=(Y_{ij})$ and soft masks $\hat{\mb{Y}}=(\hat{Y}_{ij})$, where $i=1,...,N_o$ and $ j=1,...,HWD$:
\be
\mathcal{L}_{Dice} = \sum_{k=1}^K\sum_{j=1}^{HWD} \Big[1- \frac{2Y_{kj}\hat{Y}_{kj}}{Y_{kj}+\hat{Y}_{kj}}\Big],
\ee

\be
    \mathcal{L}_{BCE} = - \sum_{k=1}^K\sum_{j=1}^{HWD} [Y_{kj} \log(\hat{Y}_{kj}) + (1-Y_{kj}) \log(1-\hat{Y}_{kj})], 
\ee
\be
    \mathcal{L} = \mathcal{L}_{Dice} + \mathcal{L}_{BCE},
\ee
where $K$ $(K \leq N_o)$ is the number of  specified masked organ tasks for every image. Note that only the loss for masked organs is calculated during each iteration, which is called the partial labeled strategy~\citep{clip-driven}.
This approach  explicitly encourages well-defined boundaries between different categories, thereby reducing the risk of mixed representations where the target and surrounding regions might be clustered together.

\subsection{Stage 2: Tumor Queries Refinement}
\begin{figure*}
	\centering
        \includegraphics[width=1\textwidth]{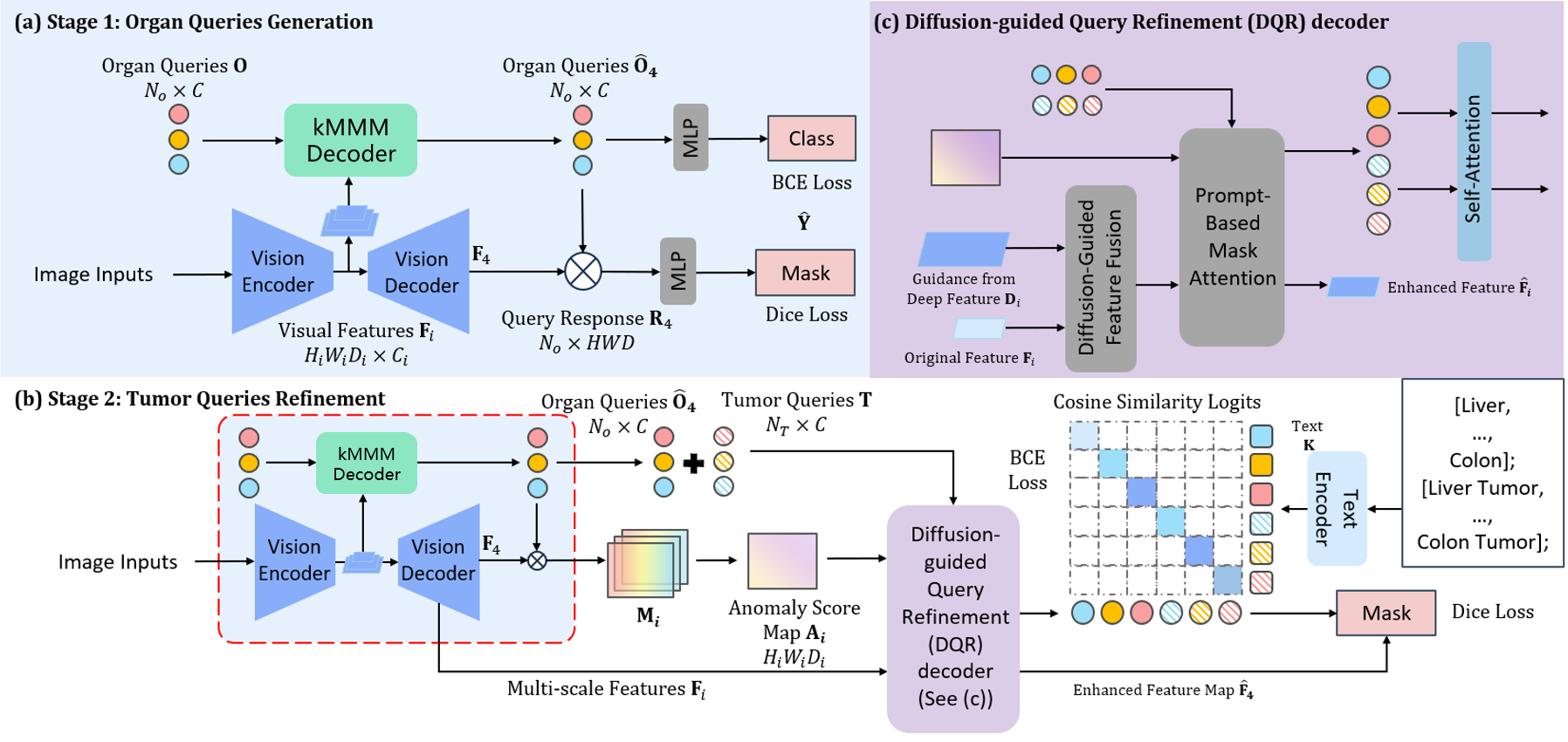}
	\caption{Pipeline overview: (a) Stage 1: Leveraging the kMaX-DeepLab \citep{yu2022k}, we introduce a feature grouping approach with SSM layers to train a series of organ queries for multi-organ segmentation. (b) Stage 2: We develop tumor queries for precise tumor segmentation by: (1) focusing on visual cues generated by OOD detection; (2) utilizing diffusion-guided boundary feature enhancement within a Transformer decoder. Finally, we integrate text embeddings with class queries to enhance the alignment, promoting cross-modal reasoning. (c) Specified explanation in Diffusion-guided Query Refinement (DQR) decoder: The multi-scale feature maps are first fused using a feature fusion strategy, then prompted with masks generated by OOD detection, and finally the  refined organ and tumor queries are output.}
	\label{overview}
\end{figure*}
After completing Stage 1, we generate a query response $\mathbf{R}_4$, indicating the affinity between visual features and organ queries, along with updated organ queries that signify different organs.
Moving on to Stage 2, our objective is to establish a new set of queries called tumor queries $\mathbf{T} \in \mathbb{R}^{N_T \times C}$ to detect and classify tumors, where $N_T$ is the total number of  tumor types. Note that $N_T$ may not necessarily match that of organ queries $N_o$, as we aim to equip the model with zero-shot segmentation capabilities.
The key idea here is to empower tumor queries  to identify anomaly information indicative of tumors by leveraging organ queries, thereby enhancing the visual delineation of tumor boundaries.

We propose a novel approach to reframe tumor segmentation as a multi-prompting process, where the tumor queries not only recognize abnormal information associated with pathological changes but also discern the accurate boundaries between organs and tumors.
To achieve this, we create mask visual prompts that enable the tumor queries to perceive abnormal information linked to pathological alterations in the features. Additionally, we implement a boundary enhancement strategy on multi-scale visual feature maps to facilitate  easier detection of tumor textures.
In Stage 1, we preserve the training datasets with organ labels to prevent the forgetting issue of the queries and introduce several datasets containing tumor labels for Stage~2. It is crucial to emphasize that the test phase encompasses novel tumor categories, validating the ``zero-shot" nature of our method.

\subsubsection{Anomaly Mask Visual Prompt}
As depicted in Figure~\ref{overview}(b), the input image and queries undergo processing by the encoder, vision decoder, and kMMM-decoder to acquire multi-scale visual features, high-resolution image embeddings, and refined organ queries, respectively. We then compute the maximum likelihood distribution embedding through the negative maximal process.
In essence, the maximal query response from outliers typically yields smaller values compared to inliers. We use the negative maximal query response $\mathbf{R}_i \in \mathbb{R}^{N_o \times H_iW_iD_i}$ within the multi-scale mask prediction $\mathbf{F}_i$ as the anomaly score map $\mathbf{A}_i$ by
\begin{align}
    \mb{A}_i = - \max_{N_o} \mb{R}_i,
    \quad i = 1,\dots, 4.
\end{align}
We then normalize the anomaly score map into mask prompts $\mathbf{M}_i$ using min-max normalization with a threshold of 0.5. For a 3D location $(h, w, d)$, the mask prompt area is
\begin{align}
    \mb{M}_i^{(h,w,d)} = \begin{cases}
                0 & \mb{A}_i^{(h,w,d)} > 0.5,\\
                -\infty & \mb{A}_i^{(h,w,d)} \leq 0.5.
    \end{cases}
\end{align}
The mask prompt $\mathbf{M}_i$ signifies the regions within the input 3D volume that pertain to an anomalous unseen category and an in-distribution seen organ class, respectively.
These mask prompts, dynamically derived from the multi-scale feature embedding space, are instrumental in guiding tumor queries to concentrate on anomalous background features and acquire proficient representations for identifying unseen tumors effectively.

\subsubsection{Diffusion-guided Query Refinement}
\begin{figure}
	\centering
        \includegraphics[width=0.4\textwidth]{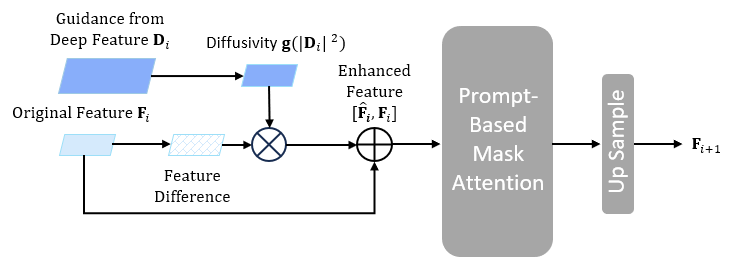}
	\caption{Diffusion-guided feature fusion mechanism.}
	\label{fig3}
\end{figure}
Inspired by the diffusion process given in Equation \eqref{eqdiff}, for multi-scale feature maps $\mathbf{F}_i$, the semantic diffusion process can be described as:
\begin{align}
    \frac{\partial \mb{F}_i}{\partial t} = Div(g(|\mb{D}_i|^2)\nabla\mb{F}_i),
\end{align}
where $\mb{D}_i$ represents the guidance feature map with the same resolution as $\mathbf{F}_i$, and $t$ refers to the time steps to enhance the guidance matrix, which in our context is $t=1$. The diffusivity term $g(|\mathbf{D}_i|^2)$ is a monotonically decreasing function of the square of the gradient.
In this context, diffusion near semantic boundary regions is dampened, while diffusion away from these boundaries is expedited. Therefore, it is imperative that $\mathbf{D}_i$ encapsulates sufficient semantic information, enabling $g(|\mathbf{D}_i|^2)$ to serve as a reliable indicator of semantic boundaries for formulating the diffusivity.

We employ a diffusion-guided feature fusion mechanism to comprehensively enhance the boundaries.
Given a feature map $\mb{F}_i$, the semantic guidance from the deep feature map $\hat{\mathbf{F}}_{i}$ is introduced to refine the original differential feature, resulting in an enhanced feature map.
The partial differential equation  governing the non-linear diffusion process can be approximately solved as  \citep{sapiro2006geometric}:
\be
    \hat{\mb{F}}_{i}[p] = \sum_{\tilde{p}\in\delta_p} g(| \mb{D}_i[\tilde p]-\mb{D}_i[p]|^2)\cdot (\mb{F}_i[\tilde p]-\mb{F}_i[p]),
\ee
\be
    \mb{F}_{i+1}[p] = {\rm Conv}(\mb{F}_i[p] +  \hat{\mb{F}}_{i}[p]),
    i = 1,...,4,
\ee
where $p$ represents the position of an entry in the feature map, and $\delta_p$ denotes the $3\times3\times3$ local region centered at $p$, $g(| \mathbf{D}_i[{\tilde{p}}] - \mathbf{D}_i[p] |^2)$ is a matrix quantifying the similarity between semantic guidance feature maps with learnable weights, and $(\mathbf{F}_i[{\tilde{p}}] - \mathbf{F}_i[{p}])$ computes the feature disparity between different positions.
As shown in Figure \ref{fig3}, Equation (17) formulates a differentiable matrix $g(|\mb{D}_i|^2)=g(| \mb{D}_i[{\tilde{p}}] - \mb{D}_i[p] |^2)$ based on the input feature map $\mb{F}_i$, transitioning to the boundary-enhanced guidance $\hat{\mb{F}}_i$ for this layer. 
Subsequently, the guidance $\hat{\mb{F}}_i$ is fused with the original feature map $\mb{F}_i$ to generate a boundary-enhanced feature, which is put into the prompt-based mask attention (see (19)) and then convolved to upsample for the next layer $\mb{F}_{i+1}$. 
By combining the differential feature map $g(|\mb{D}_i|^2)$ with the semantic guidance $\hat{\mathbf{F}}_{i}$, we derive the feature map $\hat{\mathbf{F}}_i$ with enhanced boundary details for this layer.
As illustrated in Figure \ref{overview}(c), the enhanced feature map then acts as the key and value in the prompt-based masked attention layer.

For the $i$-th block in the Diffusion-guided Query Refinement (DQR) decoder, the tumor queries are updated via interactions with enhanced feature $[\mb{F}_i,\hat{\mb{F}}_i]$ and mask prompts $\mb{A}_i$ by
\begin{align}
    \hat{\mb{T}}_i = \mb{T}_i + {\rm{Softmax}}(\mb{A}_i + \mb{Q}_{i}\mb{K}_{i}^T)\mb{V}_{i},
\end{align}
where $\mb{T}_i$ is the linear projection from $\mb{T}$ for each decoder layer, $\mb{Q}_{i}\in \mathbb{R}^{N_T \times C_i}$ is transformed from tumor query embedding $\mb{T}_i$, and $\mb{K}_{i}$ and $ \mb{V}_{i}$ are transformed from enhanced feature maps $[\mb{F}_i,\hat{\mb{F}}_i]$.
We then concatenate organ and tumor queries and apply self-attention among them to encode their relationship, facilitating the adjustment of  their representations to promote  distinct semantic differentiation between organs and tumors.
Ultimately, we utilize these updated queries to generate the corresponding mask proposals.

\subsubsection{Class Prompt Alignment}
It is common  to
use a prompt model like CLIP~\citep{clip} for zero-shot classification on the masks generated in Stage~1.
Introducing a set of fixed class prompts helps our model generalize to a broader range of unseen categories, enabling the detection and segmentation of unseen organs and tumors.
We include CLIP for weakly-supervised cross-modal alignment, aligning visual output feature maps with the high-level semantics of textual knowledge.

By utilizing CLIP as a pretrained text encoder, we generate  text embeddings 
$\{\bd{k}_i \in \mathbb{R}^C\}_{i=1}^{N_o+N_T}$
from class prompts such as ``a computerized tomography of a \{CLS\}" for every query class, where ``CLS" is a class name such as ``Liver". 
Detailed CLIP text prompts are provided below:
\begin{itemize}[leftmargin=1em]
    \item The ``CLS" for 25 organs are [`Spleen', `Right Kidney', `Left Kidney', `Gall Bladder', `Esophagus', 
                `Liver', `Stomach', `Aorta', `Postcava', `Portal Vein and Splenic Vein',
                `Pancreas', `Right Adrenal Gland', `Left Adrenal Gland', `Duodenum', `Hepatic Vessel',
                `Right Lung', `Left Lung', `Colon', `Intestine', `Rectum', 
                `Bladder', `Prostate', `Left Head of Femur', `Right Head of Femur', `Celiac Truck'].
    \item The ``CLS" for 20 tumors are [`Spleen Tumor', `Kidney Tumor', `Kidney Cyst', `Gall Bladder Tumor', `Esophagus Tumor', 
                `Liver Tumor', `Stomach Tumor', `Aortic Tumor', `Postcava Tumor Thrombus', `Portal Vein Tumor Thrombus',
                `Pancreas Tumor', `Adrenal Tumor', `Adrenal Cyst', `Duodenal Tumor', `Hepatic Vessel Tumor', 
                `Lung Tumor', `Lung Cyst', `Colon Tumor', `Small Intestinal Neoplasm', `Rectal Tumor'].
\end{itemize}
Unlike other models, we calculate the cosine similarity between the  text embeddings rather than relying solely on the vision embedding.
The predicted probability distribution for the $i$-th query is determined as:
\begin{align}
    p_i = \frac{\exp(\frac{1}{\tau}\zeta(\bd{k}_i,\bd{q}_i))}{\sum_{j=1}^{N_o+N_T}\exp(\frac{1}{\tau}\zeta(\bd{k}_j,\bd{q}_i))},
    \label{eqn: pi}
\end{align}
where $\zeta$ is the cosine similarity function, $\tau$ is the temperature parameter (set to 0.01 in our study), and $\bd{k}_i$ and $\bd{q}_i$ are the $i$-th text embedding and projected query embedding, respectively. Note that $\{\bd{q}_i\in \mathbb{R}^C\}_{i=1}^{N_o+N_T}$ is
the concatenation $[\widehat{\mb{Q}}_4^\top,\widehat{\mb{T}}_4^\top]$
after passing through a linear projection layer. During training, the similarities between matched query embedding and text embedding should be maximized. 

\subsubsection{Loss Functions}
In Stage 2, ground-truth labels and mask proposals are generalized to $\mathbb{R}^{(N_o+N_T) \times HWD}$. By adopting the cosine similarity map, we reformulate BCE loss $\mathcal{L}_{BCE}$ for semantic segmentation by 
replacing $\mb{Y}$ with $\diag(\bd{p})\mb{Y}$,
where $\bd{p}=(p_1,\dots, p_{N_o+N_T})$ with $p_i$ given in 
\eqref{eqn: pi},
 and utilize the same Dice loss  as used in Stage 1. Note that we apply the BCE loss to the cosine similarity map to enhance the knowledge alignment between our classes and the class embeddings derived from CLIP.

\begin{table}[h!]
\setlength{\tabcolsep}{3pt}
\renewcommand\arraystretch{1.1}
\centering 
\small
\scalebox{0.97}{
\begin{tabular}{l|c|c|c}
\hline
Datasets & \# Scans & Query Type & Stage\\\hline

BTCV 
& 20 & Organ & Training Stages 1\&2\\ 
CHAOS 
& 40 & Organ& Training Stages 1\&2\\
Pancreas-CT 
& 60 & Organ& Training Stages 1\&2\\
CT-ORG 
& 140 & Organ& Training Stages 1\&2\\
WORD 
& 150 & Organ& Training Stages 1\&2\\
AMOS22 
& 280 & Organ& Training Stages 1\&2\\
TotalSegmentor 
& 998 & Organ& Training Stages 1\&2\\
AbdomenCT-1K 
& 1000 & Organ& Training Stages 1\&2\\
\hline
LiTS 
& 52 & Organ \& Tumor& Training Stage 2\\
KiTS 
& 120 & Organ \& Tumor& Training Stage 2\\\hline
BTCV 
& 30  & Organ & Testing\\
LiTS 
& 79 & Organ \& Tumor& Testing\\
KiTS 
& 180 & Organ \& Tumor& Testing\\\hline
MSD CT Tasks 
& 470 & Organ \& Tumor& Inference\\\hline
\end{tabular}}

\caption{Implementation details for all datasets used.}
\label{tablescans}
\end{table}

\section{Experiments}
\subsection{Datasets}
We use CT scans from 11 publicly available datasets, including only those images with publicly released labels. 
The number of scans used from each dataset is summarized in Table \ref{tablescans}.
The data are split into training, testing, and inference sets as follows:

\begin{itemize}[leftmargin=1em]
\item {Training set (2,860 images):} In Stage 1, we use 8 public datasets (2,668 images) to train organ labels, including 
 BTCV \citep{btcv},
 CHAOS \citep{chaos}, 
 Pancreas-CT \citep{p_ct}, 
 CT-ORG \citep{ct_org}, 
 WORD \citep{word},
  AMOS22 \citep{amos22},  
  TotalSegmentor \citep{tts},
  and 
  AbdomenCT-1K \citep{abd_1k}. In Stage 2, LiTS \citep{lits} and KiTS \citep{kits} are added to the training set.
Note that 40\% of 
BTCV, LiTS, and KiTS images (192 images)
are used for training.

\item {Testing set (289 images):} 
The remaining 60\% of 
BTCV, LiTS, and KiTS images are used for testing seen tumor types.

\item {Inference set (470 images):} 
For inference on unseen tumor types, we use the MSD dataset \citep{msd}, which includes  segmentation tasks for three unseen tumor types. 
\end{itemize}

\subsection{Evaluation Metrics}
We use the Dice Similarity Coefficient (DSC) \citep{taha2015metrics} to evaluate organ and tumor segmentation performance. 
Additionally, we present the Area Under the Receiver Operating Characteristic curve (AUROC) and the False Positive Rate at a True Positive Rate of 0.95 (FPR$_{95}$) as metrics for OOD localization \citep{wu2023energy,xia2020synthesize}, given the critical importance of reducing  false positives in clinical settings.
Higher  DSC and AUROC values indicate better performance, while lower  FPR$_{95}$  values are preferred.

\subsection{Training Details}
Our DSM  is implemented in PyTorch and trained 
 on four
NVIDIA A100 GPUs.
(1) {Stage 1}: We use the Swin UNETR network~\citep{hatamizadeh2021swin} as the backbone. DSM consists of  four KMMM decoder layers ($K_d$) and four Diffusion-guided decoder layers, each processing image features with output strides of 32, 16, 8, and 4,  respectively. We employ the AdamW optimizer with a warm-up cosine scheduler spanning 50 epochs. During training, the batch size is set to 2 per GPU, with a patch size of $96 \times 96 \times 96$. The initial learning rate is $10^{-4}$, with a momentum of 0.9 and  a weight decay  of $10^{-5}$ for 500 epochs. To enhance generalization, we incorporate extensive on-the-fly data augmentation techniques,  including random rotation and scaling, elastic deformation, additive brightness, and gamma scaling. We use 25 organ queries ($N_o=25$) for  25 organs. 
(2) {Stage~2}: We utilize the pretrained model from Stage~1, while setting the initial learning rate to $4\times 10^{-4}$. The DQR decoder consists of four blocks, with each attention layer comprising eight heads. We employ 20 tumor queries ($N_T=20$) for tumors/disease categories. Other parameters are the same as in Stage 1. 

\subsection{Baseline Models}
Addressing the tasks of recognizing unseen tumors and OOD detection presents significant challenges in medical image segmentation. Hence, our baseline selection 
focuses on models with zero-shot capabilities and OOD detection. 
For inference on unseen tumor types in  MSD data,
We benchmark our DSM against (1) prominent OVSS methods, including ZegFormer~\citep{zegformer}, OpenSeg~\citep{ghiasi2022scaling}, OVSeg~\citep{OVSeg},  Freeseg~\citep{freeseg}
and ZePT~\citep{jiang2024zept}; (2) prompt-based methods specific to medical imaging, including SAMMed~\citep{wang2025sam} and Universeg~\citep{butoi2023universeg}; and (3) OOD detection methods, including SynthCP~\citep{xia2020synthesize}, SML~\citep{jung2021standardized}, and MaxQuery~\citep{yuan2023devil}.
To evaluate the ability in fully-supervised training, we compare our DSM against medical image segmentation models, including nnUNet \citep{isensee2021nnu}, the Universal model \citep{clip-driven}, Swin UNETR \citep{hatamizadeh2021swin}, and ZePT \citep{jiang2024zept}
on  testing and inference datasets.
We use the pretrained weights provided by the baseline models, except for Swin UNETR and ZePT, which are fine-tuned on the inference dataset since they were not originally pretrained on this data.

\begin{table*}[p]
\setlength{\tabcolsep}{3.5pt}
\renewcommand\arraystretch{1.2}
\centering 
\scalebox{0.97}{
\begin{tabular}{l|ccc|ccc|ccc}
\hline
\multicolumn{1}{c|}{\multirow{2}{*}{Method}}
& \multicolumn{3}{c|}{Pancreas Tumor} & \multicolumn{3}{c|}{Lung Tumor}&\multicolumn{3}{c}{Colon Tumor} \\ \cline{2-10}
\multicolumn{1}{c|}{}                        
& \multicolumn{1}{c}{AUROC$\uparrow$} & \multicolumn{1}{c}{FPR$_{95}\downarrow$} & \multicolumn{1}{c|}{DSC$\uparrow$} 
& \multicolumn{1}{c}{AUROC$\uparrow$} & \multicolumn{1}{c}{FPR$_{95}\downarrow$} & \multicolumn{1}{c|}{DSC$\uparrow$} 
& \multicolumn{1}{c}{AUROC$\uparrow$} & \multicolumn{1}{c}{FPR$_{95}\downarrow$} & \multicolumn{1}{c}{DSC$\uparrow$} 
\\ \cline{1-10}
ZegFormer 
& $66.45{\pm 1.14}$ &$ 69.33{\pm 1.28}$ & $14.92{\pm 0.76}$ 
& $41.31{\pm 0.36}$ & $81.78{\pm 4.33}$ & $9.94{\pm 0.49}$  
&$50.13{\pm 0.07}$&$78.81{\pm 0.07}$&$11.34{\pm 0.16}$ \\

OpenSeg 
& $44.56{\pm 1.34}$&$85.19{\pm 0.91}$&$10.05{\pm 0.34}$
& $23.49{\pm 0.72}$&$91.75{\pm 6.31}$&$6.12{\pm 0.59}$
& $41.76{\pm 0.10}$&$89.44{\pm 0.05}$&$7.13{\pm 0.15}$ \\

OVSeg 
& $70.22{\pm 0.71}$&$59.73{\pm 1.05}$&$19.36{\pm 0.43}$
&$52.93{\pm 0.46}$&$68.65{\pm 0.53}$&$14.11{\pm 0.55}$ 
&$59.94{\pm 0.07}$&$65.25{\pm 0.08}$&$15.76{\pm 0.14}$ \\

Freeseg 
& $69.98{\pm 0.75}$&$60.75{\pm 0.96}$&$18.19{\pm 0.44}$
&$49.92{\pm 0.32}$&$70.39{\pm 0.43}$&$13.26{\pm 0.43}$ 
&$56.45{\pm 0.05}$&$68.49{\pm 0.04}$&$14.71{\pm 0.13}$ \\

ZePT 
& $86.81{\pm 0.65}$&$35.18{\pm 0.82}$&$37.67{\pm 0.29}$
&$\mathbf{77.84{\pm 0.32}}$&$44.30{\pm 1.02}$&$27.22{\pm 0.21}$
&$82.36{\pm 0.04}$&$40.73{\pm 0.03}$&$30.45{\pm 0.05}$\\ 
\hline
SAMMed 
& $72.75{\pm 1.24}$&$57.84{\pm 1.43}$&$22.43{\pm 0.23}$
&$64.28{\pm 0.28}$&$48.27{\pm 2.99}$&$17.97{\pm 0.23}$
&$61.47{\pm 0.14}$&$63.55{\pm 0.05}$&$16.46{\pm 0.11}$\\

Universeg 
& $79.64{\pm 1.05}$&$40.49{\pm 1.01}$&$32.51{\pm 0.50}$
&$67.22{\pm 0.37}$&$48.19{\pm 1.21}$&$23.38{\pm 0.12}$ 
&$79.29{\pm 0.09}$&$45.51{\pm 0.03}$&$28.72{\pm0.05}$\\ \hline

SynthCP 
& $51.24{\pm 1.42}$&$81.69{\pm 1.45}$&$11.33{\pm 0.25}$
&$25.85{\pm 0.41}$&$90.28{\pm 5.13}$&$6.43{\pm 0.32}$
&$43.84{\pm 0.15}$&$87.71{\pm 0.07}$&$8.74{\pm 0.07}$\\

SML 
& $37.95{\pm 1.90}$&$89.93{\pm 1.15}$&$9.72{\pm 0.18}$
&$20.18{\pm 0.39}$&$93.65{\pm 4.91}$&$6.02{\pm 0.39}$ 
&$22.41{\pm 0.11}$&$92.07{\pm 0.15}$&$6.65{\pm 0.09}$ \\

MaxQuery 
& $68.99{\pm 1.03}$&$59.93{\pm 1.11}$&$18.15{\pm 0.70}$
&$48.24{\pm 0.57}$&$70.47{\pm 4.40}$&$11.29{\pm 0.32}$ 
&$50.47{\pm 0.09}$&$69.88{\pm 0.03}$&$13.43{\pm0.05}$\\ \hline

DSM 
& $\mathbf{88.61{\pm 0.65}}$ & $\mathbf{30.04{\pm 0.81}}$ & $\mathbf{38.86{\pm 0.26}}$ 
& $76.91{\pm 0.35}$ & $\mathbf{43.72{\pm 1.01}}$ & $\mathbf{28.65{\pm 0.32}}$
& $\mathbf{83.34{\pm 0.04}}$ & $\mathbf{41.78{\pm 0.04}}$ & $\mathbf{31.52{\pm 0.05}}$ \\ \hline
\end{tabular}}
\caption{Segmentation performance of unseen tumors on MSD dataset. All scores (mean $\pm$ SE) are in \%.}
\label{table1}
\end{table*}

\begin{table*}[p]
\setlength{\tabcolsep}{3.5pt}
\renewcommand\arraystretch{1.1}
\footnotesize
\centering 
\begin{tabular}{l|ccc|ccc|ccc}
\hline
\multicolumn{1}{c|}{\multirow{2}{*}{Method}}
& \multicolumn{3}{c|}{BTCV} & \multicolumn{3}{c|}{LiTS}& \multicolumn{3}{c}{KiTS}\\ \cline{2-10}
\multicolumn{1}{c|}{}                        
& \multicolumn{1}{c}{AUROC$\uparrow$} & \multicolumn{1}{c}{FPR$_{95}\downarrow$} & \multicolumn{1}{c|}{DSC$\uparrow$} 
& \multicolumn{1}{c}{AUROC$\uparrow$} & \multicolumn{1}{c}{FPR$_{95}\downarrow$} & \multicolumn{1}{c|}{DSC$\uparrow$} 
& \multicolumn{1}{c}{AUROC$\uparrow$} & \multicolumn{1}{c}{FPR$_{95}\downarrow$} & \multicolumn{1}{c}{DSC$\uparrow$} 
\\ \cline{1-10}
nnUNet 
& ${88.07\pm0.17}$ &${24.85\pm0.15}$&$82.23\pm0.30$ &${91.21\pm0.12}$&$21.03{\pm0.15}$&  $77.15\pm0.30$  &${93.35{\pm0.10}}$&$18.76{\pm0.11}$&  $85.18\pm0.09$ \\ 

Universal 
&${94.10\pm0.15}$&${19.77\pm0.11}$& $86.38\pm0.23$ &${92.88\pm0.09}$&$19.72{\pm0.12}$& $80.58\pm0.26$ &${95.06{\pm0.06}}$&$16.18{\pm0.07}$& $87.05\pm0.08$ \\ 

Swin UNETR 
&${90.01\pm0.16}$&${21.31\pm0.14}$& $82.26\pm0.29$ &${90.01\pm0.12}$&$21.11{\pm0.14}$& $76.79\pm0.31$ &${94.45{\pm0.10}}$&$18.22{\pm0.10}$& $85.52\pm0.08$  \\

ZePT 
&${96.75\pm0.12}$&${17.04\pm0.06}$& $87.09\pm0.22$ &${93.03\pm0.08}$&$\mathbf{18.07{\pm0.09}}$& $81.66\pm0.24$ &${95.97{\pm0.06}}$&$15.07{\pm0.05}$& $87.73\pm0.07$ \\ \hline
DSM &$\mathbf{96.87\pm0.11}$&$\mathbf{16.88\pm0.06}$& $\mathbf{88.24\pm0.21}$ &$\mathbf{93.24{\pm0.08}}$&$18.56{\pm0.11}$& $\mathbf{82.73\pm0.24}$ &$\mathbf{96.02{\pm0.05}}$&$\mathbf{14.98{\pm0.05}}$& $\mathbf{88.62\pm0.07}$\\\hline
\end{tabular}
\caption{Segmentation results for 13 organs in BTCV, liver tumor in LiTS, and kidney tumor in KiTS. All scores
(mean $\pm$ SE) are in \%. }
\label{tableblk}
\end{table*}

\begin{table*}[p]
\setlength{\tabcolsep}{3.5pt}
\renewcommand\arraystretch{1.2}
\centering 
\scalebox{0.88}{
\begin{tabular}{l|ccc|ccc|ccc}
\hline
\multicolumn{1}{c|}{\multirow{2}{*}{Method}}
& \multicolumn{3}{c|}{Pancreas Tumor} & \multicolumn{3}{c|}{Lung Tumor}&\multicolumn{3}{c}{Colon Tumor} \\ \cline{2-10}
\multicolumn{1}{c|}{}                        
& \multicolumn{1}{c}{AUROC$\uparrow$} & \multicolumn{1}{c}{FPR$_{95}\downarrow$} & \multicolumn{1}{c|}{DSC$\uparrow$} 
& \multicolumn{1}{c}{AUROC$\uparrow$} & \multicolumn{1}{c}{FPR$_{95}\downarrow$} & \multicolumn{1}{c|}{DSC$\uparrow$} 
& \multicolumn{1}{c}{AUROC$\uparrow$} & \multicolumn{1}{c}{FPR$_{95}\downarrow$} & \multicolumn{1}{c}{DSC$\uparrow$} 
\\ \cline{1-10}
nnUNet 
& ${91.75{\pm 0.07}}$ & $25.88{\pm 0.09}$ & $50.45{\pm 0.13}$
& $90.75{\pm 0.06}$ & $24.56{\pm 0.16}$ & $66.57{\pm 0.18}$
& $91.55{\pm 0.05}$ & $23.75{\pm 0.07}$ & $50.07{\pm 0.07}$ \\

Universal
& ${93.32{\pm 0.05}}$ & $19.12{\pm 0.06}$ & $60.83{\pm 0.10}$ 
& $94.53{\pm 0.04}$ & $18.85{\pm 0.14}$ & $67.11{\pm 0.15}$ 
& $93.15{\pm 0.05}$ & $19.24{\pm 0.05}$ & $62.15{\pm 0.06}$\\ 

Swin UNETR (fine-tuned) 
& ${91.83{\pm 0.06}}$ & $24.47{\pm 0.08}$ & $52.54{\pm 0.12}$ 
& $92.01{\pm 0.05}$ & $22.31{\pm 0.16}$ & $68.90{\pm 0.16}$ 
& $91.73{\pm 0.05}$ & $22.34{\pm 0.08}$  & $50.55{\pm 0.06}$ \\

ZePT (fine-tuned) 
& $93.85{\pm 0.05}$ & $18.87{\pm 0.07}$ & $62.10{\pm 0.10}$ 
& $93.97{\pm 0.05}$ & $18.17{\pm 0.12}$ & $62.15{\pm 0.16}$ 
& $\mathbf{93.97{\pm 0.05}}$ & $18.91{\pm 0.05}$ & $\mathbf{64.87{\pm 0.05}}$ \\\hline

DSM (fine-tuned) 
& $\mathbf{94.07{\pm 0.04}}$ & $\mathbf{18.04{\pm 0.11}}$ & $\mathbf{62.25{\pm 0.09}}$ 
& $\mathbf{94.91{\pm 0.04}}$ & $\mathbf{17.72{\pm 0.12}}$ & $\mathbf{69.45{\pm 0.14}}$
& ${93.52{\pm 0.04}}$ & $\mathbf{18.68{\pm 0.04}}$ & $63.31{\pm 0.05}$ \\ \hline
\end{tabular}
}
\caption{Results for fully supervised segmentation  on MSD dataset. All scores (mean $\pm$ SE) are in \%.}
\label{tablefull}
\end{table*}

\begin{table*}[p]
\setlength{\tabcolsep}{3.5pt}
\renewcommand\arraystretch{1.2}
\centering 
\scalebox{0.9}{
\begin{tabular}{cccc|ccc|ccc}
\hline
\multicolumn{4}{c|}{\multirow{2}{*}{Modules}} & \multicolumn{3}{c|}{Liver Tumor}    & \multicolumn{3}{c}{Colon Tumor}       \\ 
\multicolumn{4}{c|}{}       & \multicolumn{3}{c|}{in LiTS (Seen)}                     & \multicolumn{3}{c}{in MSD (Unseen)}   \\ \cline{1-10}
kMMM&AMVP&DQR&CLIP       & \multicolumn{1}{c}{AUROC$\uparrow$}& \multicolumn{1}{c}{FPR$_{95}$ $\downarrow$} & \multicolumn{1}{c|}{DSC$\uparrow$} & \multicolumn{1}{c}{AUROC$\uparrow$} & \multicolumn{1}{c}{FPR$_{95}$ $\downarrow$} & DSC$\uparrow$   \\ \cline{1-10}
 & & & &$90.92{\pm0.14}$ & $23.18{\pm0.15}$ & $78.98{\pm0.28}$ & $78.22{\pm0.05}$ & ${45.82{\pm 0.08}}$ & $25.07{\pm0.07}$ \\
&$\checkmark$ & $\checkmark$&$\checkmark$ &$92.23{\pm0.13}$ & $21.47{\pm0.15}$ & $80.22{\pm0.26}$ & $79.18{\pm0.05}$& ${43.77{\pm 0.07}}$ & $27.31{\pm0.07}$ \\
$\checkmark$ &  &$\checkmark$ &$\checkmark$& $93.14{\pm0.11}$& $21.11{\pm0.13}$ & $81.16{\pm0.27}$ & $79.03{\pm0.05}$& ${42.64{\pm 0.05}}$ & $26.84{\pm0.06}$ \\
$\checkmark$ &  $\checkmark$& &$\checkmark$& $92.73{\pm0.09}$& $19.41{\pm0.12}$ & $80.84{\pm0.25}$ & $80.64{\pm0.04}$& ${43.75{\pm 0.05}}$ & $29.03{\pm0.06}$  \\
$\checkmark$ & $\checkmark$ & $\checkmark$ && $91.73{\pm0.11}$ & $20.03{\pm0.13}$ & $80.65{\pm0.27}$ & $79.32{\pm0.05}$ & ${43.45{\pm 0.08}}$ & $29.76{\pm0.07}$ \\
$\checkmark$ & $\checkmark$ & $\checkmark$ & $\checkmark$&  $\mathbf{93.24{\pm0.08}}$&  $\mathbf{18.56{\pm0.11}}$ &  $\mathbf{82.73{\pm0.24}}$ &  $\mathbf{83.34{\pm0.04}}$&  $\mathbf{41.78{\pm 0.04}}$ &  $\mathbf{31.52{\pm0.05}}$ \\\hline
\end{tabular}}

\caption{Ablation study of DSM's different components on LiTS and MSD datasets. All  scores
(mean $\pm$ SE) are in \%.}
\label{tableablation}
\end{table*}

\section{Results}

\subsection{Unseen Tumor Segmentation on MSD Data} 

We evaluate DSM on unseen tumors, with results summarized in Table~\ref{table1}. Across three unseen tumor categories  from MSD, our DSM 
 achieves the best performance  in 8 out of  9 evaluation terms,
and ranks  second in the remaining one with a result  comparable to the top score.

Compared to the state-of-the-art (SOTA) OVSS methods, 
DSM exhibits superior performance,
with the only exception being a slightly lower mean AUROC than ZePT in  lung tumor segmentation,
underscoring DSM's strong ability to localize unseen tumors.
These results demonstrate  that the proposed diffusion-guided boundary enhancement and multiple prompting mechanisms effectively link visual features to query embeddings, ultimately enhancing the model's  localization performance on unseen tumors.

In comparison to SAMMed and Universeg,
two prompt-based models designed for medical image segmentation, our  DSM consistently outperforms both  across all tumor categories and evaluation metrics, highlighting its strong generalizability  for unseen tumor segmentation. While UniverSeg significantly improves upon SAMMed, particularly in  AUROC, it still lags behind DSM in segmenting less represented tumor types.

Relative to  SOTA OOD detection methods, DSM surpasses the  best-performing baseline, MaxQuery, across three tasks  by averagely at least
0.1962 in mean AUROC, 
0.2675 in mean FPR$_{95}$, and 0.1736 in mean DSC. Unlike most OOD region segmentation methods that solely exploit visual modality information, DSM aligns visual features with linguistic semantics for cross-modal interaction. The results suggest that combining image features with medical domain knowledge enhances the semantic understanding of unseen tumors.

Figure~\ref{fig1} visualizes segmentation results on the MSD dataset from DSM and those baseline methods, where DSM demonstrates the best segmentation performance.

\subsection{Segmentation of Seen Organs and Tumors on Testing Data}

We evaluate the segmentation performance of DSM on familiar organs using the testing data and compare it with the four baselines (nnUNet, the Universal model, Swin UNETR, and ZePT). 
Table \ref{tableblk} shows that DSM outperforms all other methods, with an average improvement of at least 0.0115 in DSC for BTCV, 0.0107 for liver tumors in LiTS, and 0.0089 for kidney tumors in KiTS. Notably, 
the Universal model \citep{clip-driven} and ZePT \citep{jiang2024zept} adopt a CLIP text encoder and design an architecture for query refinement but use the original visual features. 
DSM takes one step further by designing a diffusion-guided boundary enhancement structure on visual features to improve the semantic segmentation.
These improvements demonstrate that DSM can also segment seen organs and tumors with high accuracy.

\subsection{Fully Supervised Segmentation on MSD Data}

To evaluate the fully supervised performance of DSM on the MSD dataset, we fine-tune it on this dataset and compare it with the other four SOTA segmentation models. The MSD dataset we used, consisting of 470 CT scans for Lung, Colon, and Pancreas, is split into 396 images for training and 74 for testing. We apply the same data split to fine-tune Swin UNETR and ZePT. For nnUNet and the Universal model, which were pretrained on the MSD dataset, we use their pretrained models directly for comparison. The results are presented in Table \ref{tablefull}.
Our DSM achieves the best performance in 7 out of 9 evaluation terms. In the remaining two, it ranks second, with results that are comparable to the best.

\begin{figure*}[t!]
	\centering
        \includegraphics[width=1\textwidth]{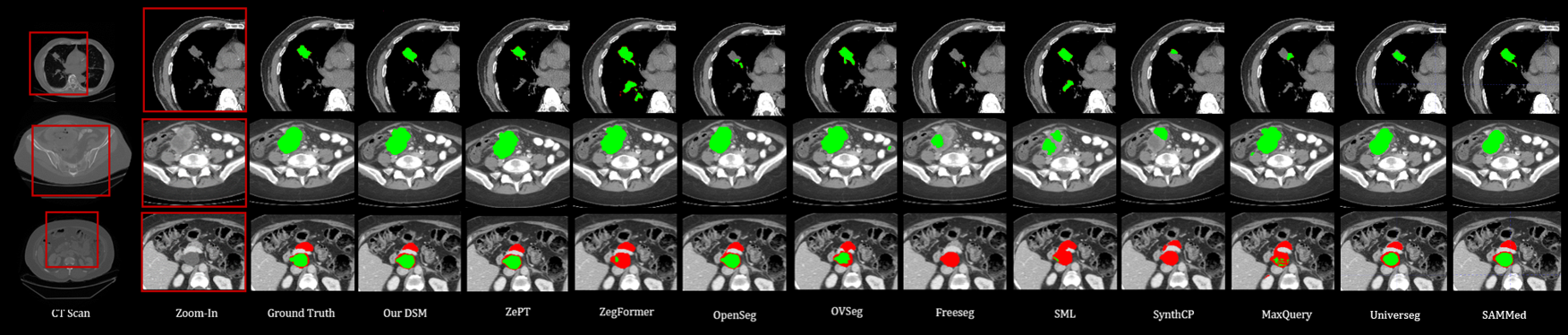}
	\caption{Qualitative visualization of the proposed DSM compared with competing models 
    in unseen tumor segmentation
    on MSD dataset. The segmentation results shown from Rows 1 to 3 correspond to Lung Tumor, Colon Tumor, and Pancreas Tumor, respectively. The red region represents the organ and the green region represents the tumor.}
	\label{fig1}

 \bigskip
 \includegraphics[width=0.75\textwidth]{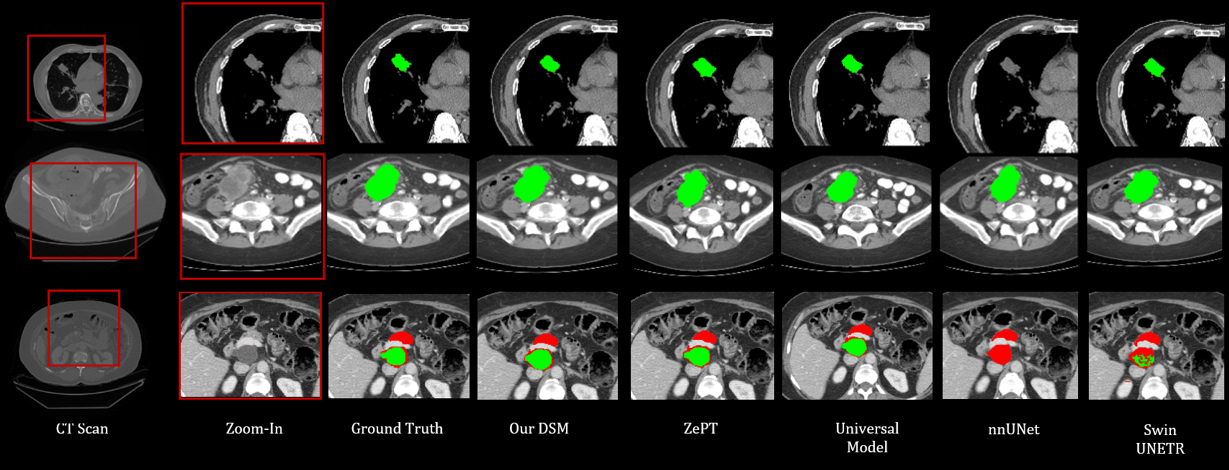}
\caption{Qualitative visualization of the proposed DSM compared with competing models in fully supervised segmentation on MSD dataset. The segmentation results shown from Rows 1 to 3 correspond to Lung Tumor, Colon Tumor, and Pancreas Tumor, respectively.
    The red region represents the organ and the green region represents the tumor.}
	\label{Qualitative result}
\end{figure*}

Figure \ref{Qualitative result} shows the qualitative results on DSM with medical segmentation models and highlights the advantages of DSM. Most competing methods suffer from incomplete segmentation of the target or misclassification of background regions as tumors (false positives). In contrast, DSM produces sharper boundaries and generates results that are more consistent with the ground truth than  all other models.


\subsection{Ablation Studies}

We conduct ablation studies
to evaluate the effectiveness of four key components
in our DSM framework, including the kMMM block,  
the anomaly mask visual prompt attention
(AMVP), the DQR process, and the CLIP text embedding.
The evaluation is performed on liver tumor segmentation in the LiTS dataset and colon tumor segmentation in the MSD dataset. Results summarized in Table~\ref{tableablation} show that the full model incorporating all four components achieves the best performance.

{\bf Significance of kMMM.}
Rows 2 and 6 of Table~\ref{tableablation} compare the kMMM blocks with the vanilla MaskFormer blocks  for query updates with the Swin UNETR backbone.  A modest performance improvement  is observed with  kMMM on both the testing and inference sets, indicating that incorporating  an SSM is  beneficial for maintaining strong performance during long-sequence modeling tasks.

{\bf Efficiency of Mask Prompts in OOD.}
As shown in Rows~3 and 6 of Table \ref{tableablation}, the AMVP is compared with a vanilla self-attention layer, directly concatenating multiple anomaly score maps with image features to update the advanced queries. Removing AMVP leads to a decline in performance for both seen and unseen tumors.

{\bf Importance of Diffusion-Guided Query Refinement.} 
As depicted in Rows 4 and 6 of Table \ref{tableablation}, the DQR process is compared  with an alternative approach using a projection layer. The feature map undergoes processing through the projection layer and is fused with the upper layer feature map. The result in Table \ref{tableablation} suggests that feature fusion and boundary enhancement play a significant role in semantic segmentation tasks. Various methods introduce diffusion process into segmentation networks to enhance performance in diverse ways \citep{tan2022semantic, wang2023dformer}, and we believe this is a general approach that can benefit deep learning models in various popular
visual tasks.

{\bf Usage of CLIP text embedding.}
As shown in Rows 5 and 6 of Table \ref{tableablation}, using the cosine similarity map with CLIP text embedding outperforms the approach of using the Hungarian matching algorithm, which is used in  MaskFormer \citep{maskformer,mask2former} to match queries with  mask proposals. 

{\bf Model without  the 4 Components.}
Row 1 of Table \ref{tableablation} shows the result for the model in which all the four components are replaced by cross-attention layers. This model demonstrates the poorest performance among all the models presented in Table~\ref{tableablation}.

\section{Conclusion}
In this paper,
we propose DSM, a pioneering segmentation framework that leverages query refinement and visual boundary enhancement for precise semantic organ segmentation and tumor detection. A key innovation is the incorporation of kMMM layers to bolster long-term memory within query embeddings. We also introduce a mask visual prompt to prioritize tumor detection and utilize the diffusion-guided query decoder for refined tumor boundary enhancement. Furthermore, we enhance the model's robustness by conducting cosine similarity alignment between class and text embeddings.
The substantial performance improvements observed with DSM across multiple organ and tumor segmentation tasks underscore its efficacy.

\section*{Acknowledgements}\label{sec:acknowledgement}
This research was partially supported by Dr. Hai Shu's NYU GPH Research Support Grant.
 This work was supported in part through the NYU IT High Performance Computing resources, services, and staff expertise.

\bibliographystyle{elsarticle-harv} 
\bibliography{main}


\end{document}